\documentclass[11pt,a4paper]{article}
\usepackage[hyperref]{eacl2021}
\usepackage{times}
\usepackage{latexsym}
\usepackage{epsfig}
\usepackage{graphicx}
\usepackage{xcolor}
\usepackage{url}

\usepackage{amsmath}
\usepackage{amssymb}

\usepackage{microtype}

\aclfinalcopy 
\def\aclpaperid{362} 

\title{Clustering Word Embeddings with Self-Organizing Maps. Application on \textcolor{red}{La}\textcolor{yellow}{Ro}\textcolor{blue}{Se}\textcolor{green}{Da} -- A \textcolor{red}{La}rge \textcolor{yellow}{Ro}manian \textcolor{blue}{Se}ntiment \textcolor{green}{Da}ta Set}

\author{Anca Maria Tache$^1$, Mihaela G\u{a}man$^1$, Radu Tudor Ionescu$^{1,2,*}$\\
  $^1$Department of Computer Science, $^2$Romanian Young Academy\\
  University of Bucharest\\
  14 Academiei, Bucharest, Romania\\
  $^*$\texttt{raducu.ionescu@gmail.com} \\}

\date{}

\begin{document}
\maketitle
\begin{abstract}
Romanian is one of the understudied languages in computational linguistics, with few resources available for the development of natural language processing tools. In this paper, we introduce LaRoSeDa, a \textbf{La}rge \textbf{Ro}manian \textbf{Se}ntiment \textbf{Da}ta Set\footnote{https://github.com/ancatache/LaRoSeDa}, which is composed of 15,000 positive and negative reviews collected from one of the largest Romanian e-commerce platforms. We employ two sentiment classification methods as baselines for our new data set, one based on low-level features (character n-grams) and one based on high-level features (bag-of-word-embeddings generated by clustering word embeddings with k-means). As an additional contribution, we replace the k-means clustering algorithm with self-organizing maps (SOMs), obtaining better results because the generated clusters of word embeddings are closer to the Zipf's law distribution, which is known to govern natural language. We also demonstrate the generalization capacity of using SOMs for the clustering of word embeddings on another recently-introduced Romanian data set, for text categorization by topic.
\end{abstract}

\setlength{\abovedisplayskip}{4pt}
\setlength{\belowdisplayskip}{4pt}

\section{Introduction}

Perhaps one of the most studied tasks in computational linguistics is sentiment classification, a.k.a.~opinion mining or polarity classification. The task has been studied across several languages, the most popular being English \cite{Blitzer-ACL-2007,Dos-COLING-2014,Fu-ESA-2018,Gimenez-EACL-2017,Huang-AAAI-2017,Ionescu-NAACL-2019,Kim-EMNLP-2014,Maas-ACL-2011,Pang-ACL-2005,Shen-ACL-2018,Socher-EMNLP-2013}, Chinese \cite{Peng-CC-2017,Wan-EMNLP-2008,Zagibalov-IJCNLP-2008,Zhai-ESA-2011,Zhang-JIS-2013,Zhang-ICCIT-2008}, Arabic \cite{Al-Ayyoub-JCS-2018,Elsahar-CICLing-2015,Dahou-COLING-2016,Nabil-ACL-2013,Nabil-EMNLP-2015} or Spanish \cite{Brooke-RANLP-2009,Molina-IPM-2015,Navas-LRE-2019,Vilares-NLE-2015,Zafra-TAC-2017}. However, studying this task for under-studied languages, e.g. Romanian, is difficult without access to large data sets.
We hereby introduce LaRoSeDa, a \textbf{La}rge \textbf{Ro}manian \textbf{Se}ntiment \textbf{Da}ta Set, which is freely available for download at {https://github.com/ancatache/LaRoSeDa} for noncommercial use. With a total of 15,000 positive and negative reviews collected from one of the most popular Romanian e-commerce websites, to our knowledge, LaRoSeDa is the largest data set for Romanian polarity classification.

We experiment with two baseline methods on our novel data set. The first baseline employs string kernels, an approach based on low-level features (character n-grams), that was found to work well for sentiment analysis across multiple languages, e.g.~English \cite{Gimenez-EACL-2017,Ionescu-EMNLP-2018}, Chinese \cite{Zhang-ICCIT-2008} and Arabic \cite{Popescu-KES-2017}, requiring no linguistic resources besides a labeled training set of samples. The second baseline employs bag-of-word-embeddings \cite{Ionescu-NAACL-2019,Fu-ESA-2018}, an approach based on high-level features (clusters of word embeddings generated by k-means), that attains good results in various text classification tasks \cite{Ionescu-KES-2017,Cozma-ACL-2018,Fu-ESA-2018,Ionescu-NAACL-2019}, including sentiment analysis. As an additional contribution, we replace the k-means clustering algorithm in the second baseline method with self-organizing maps (SOMs) \cite{Kohonen-Springer-2001}, obtaining better results because the generated clusters of word embeddings are closer to the Zipf's law distribution (see Figure~\ref{fig2}), which is known to govern natural language \cite{Powers-CoNLL-1998}. To our knowledge, we are the first to apply SOMs to cluster word embeddings, showing performance gains for both word2vec~\cite{Mikolov-NIPS-2013} and Romanian BERT~\cite{Dumitrescu-EMNLP-2020} embeddings. We also demonstrate the generalization capacity of using SOMs in the bag-of-word-embeddings on another recently-introduced Romanian data set \cite{Butnaru-ACL-2019}, for the task of text categorization by topic.

In summary, our contribution is twofold:
\begin{itemize}
    \item \vspace{-0.2cm} We introduce LaRoSeDa, one of the largest corpora for Romanian sentiment analysis, along with a set of strong baselines to be used as reference in future research.
    \item \vspace{-0.2cm} To our knowledge, we are the first to employ SOMs as a technique to cluster word embeddings. We provide empirical evidence showing that SOMs produce better results than the popular k-means.
\end{itemize}

\section{Related Work}

To date, a small number of works targeting sentiment classification in the Romanian language have been published. Preceding the sentiment analysis efforts on Romanian texts, there are a few studies on subjectivity, that have introduced two corpora built through cross-lingual projections from English to Romanian \cite{Mihalcea-ACL-2007} or through machine translation \cite{Banea-EMNLP-2008}. An extensive study conducted by \newcite{Banea-BC-2011} looks at sentiment and subjectivity from a computational linguistics perspective, in a multilingual setup in which Romanian is also included. However, in these initial works, Romanian is studied only from a subjectivity perspective, which does not go down to the level of polarity.

\begin{figure}[!t]
\begin{center}
\hspace{-0.4cm}
\includegraphics[width=0.82\linewidth]{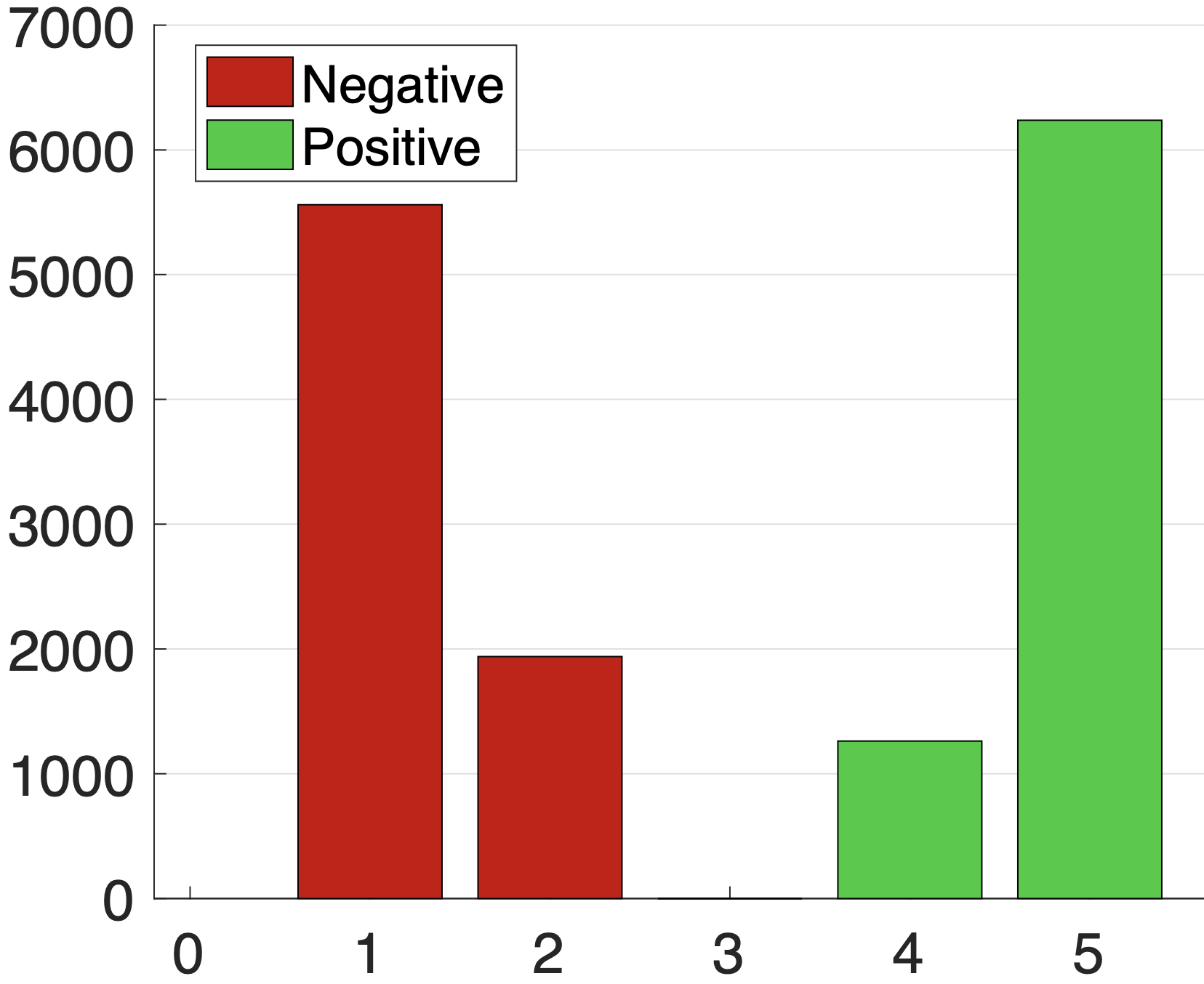}
\end{center}
\vspace*{-0.3cm}
\caption{The rating distribution of Romanian product reviews. Negative reviews are those rated with one or two stars, while positive reviews are those rated with four or five stars. Neutral reviews are not included in our data set. Best viewed in color.}
\label{fig1}
\vspace*{-0.4cm}
\end{figure}

On our topic (i.e.~sentiment analysis in Romanian), the first study that we have found describes two word sets tagged with polarity for Romanian and Russian \cite{Sokolova-RANLP-2009}. \newcite{Ginsca-WASSA-2011} introduced a sentiment analysis service intended for multiple languages, that also supports Romanian. They perform sentiment identification using a list of  manually-built triggers which, to our knowledge, is not publicly available. Another effort \cite{Colhon-INISTA-2016} in creating an opinion lexicon with polarity annotations introduced a collection of 2,521 Romanian tourist reviews and an extensive linguistic analysis of the corpus. The data set is not released for public use. Similarly, we did not find any public link to RoEmoLex, a lexicon with approximately 11,000 Romanian words tagged for emotion and sentiment \cite{Briciu-SUBBI-2017, Lupea-ICCP-2017}.
The only Romanian data set annotated for sentiment that we have found freely available is rather small, with 1,000 movie reviews manually extracted from several blogs and sites \cite{Russu-ICCP-2014}. With 15,000 reviews, our corpus is much larger.

\section{Data Set}

In order to build LaRoSeDa, we collected product reviews from one of the largest e-commerce websites in Romania. Along with the textual content of each review, we collected the associated star ratings in order to automatically assign labels to the collected text samples. Following the same approach used for data sets containing English reviews \cite{Blitzer-ACL-2007,Maas-ACL-2011,Pang-ACL-2005}, we assigned positive labels to the reviews rated with four or five stars and negative labels to the reviews rated with one or two stars. However, the star rating might not always reflect the polarity of the text. We thus acknowledge that the automatic labeling process is not optimal, i.e.~some labels might be noisy. Since automatic labeling based on star ratings is a commonly accepted practice for opinion mining data sets of product reviews, we leave the analysis of noisy labels and manual labeling for future work.
\begin{table*}[!t]
\begin{center}
\begin{tabular}{|l|r|r|r|r|}
\hline
Set 						& \multicolumn{2}{|c|}{Positive}    & \multicolumn{2}{|c|}{Negative}\\
\cline{2-5}
     						& \#samples		&	\#words	    & \#samples		&	\#words\\

\hline
\hline
Training				    & 6,000 		& 187,813            & 6,000 		& 244,499\\
Test					    & 1,500			& 47,661             & 1,500		& 60,314\\
\hline
Total						& 7,500			& 235,474           & 7,500			& 304,813\\
\hline
\end{tabular}
\end{center}
\vspace*{-0.2cm}
\caption{The number of positive and negative samples (\#samples) and the corresponding number of words (\#words) contained in the training set and the test set of LaRoSeDa.}
\label{tab_stats}
\vspace*{-0.2cm}
\end{table*}

We also imitate the data collection approach used for English review data sets \cite{Blitzer-ACL-2007,Maas-ACL-2011,Pang-ACL-2005}, selecting a balanced set of Romanian reviews. More precisely, LaRoSeDa is formed of a total of 15,000 reviews that are perfectly balanced, i.e.~half of them (7,500) are positive reviews and the other half (7,500) are negative reviews.
In Figure~\ref{fig1}, we show the distribution of reviews with respect to the star ratings. We note that most of the negative reviews (5,561) are rated with one star. Similarly, most of the positive reviews (6,238) are rated with five stars. Hence, the corpus is highly polarized. 
We divide LaRoSeDa into a training set containing 80\% of the data samples and a test set containing the remaining 20\%. In Table~\ref{tab_stats}, we present the number of positive and negative reviews inside each subset, along with the number of words. Our data set contains a total of 540,287 words, with an average of 36 words per review. We observe that positive reviews contain 235,474 words (44.6\%) and negative reviews contain 304,813 words (56.4\%). We note that, in negative reviews, people are likely to complain about several points or to explain what is wrong with the reviewed products. This could provide a natural explanation for the fact that the negative reviews contain more words than the positive reviews.

\section{Methods}

\noindent
{\bf String kernels.}
A simple language-independent and linguistic-theory-neutral approach is to interpret text samples as sequences of characters (strings) and to use character n-grams as features. The number of character n-grams is usually much higher than the number of samples, so representing the text samples as feature vectors may require a lot of space. String kernels provide an efficient way to avoid storing and using the feature vectors (primal form), by representing the data though a kernel matrix (dual form). Each component $K_{ij}$ in a kernel matrix represents the similarity between data samples $x_i$ and $x_j$. In our experiments, we use the histogram intersection string kernel (HISK) \cite{Ionescu-EMNLP-2014,Ionescu-COLI-2016} as the similarity function. For two strings $x_i$ and $x_j$ over a set of characters $S$, HISK is defined as follows:
\begin{equation}
\begin{split}
k^{\cap}(x_i, x_j)=\sum\limits_{g \in S^n} \min \lbrace \mbox{\#}(x_i, g), \mbox{\#}(x_j, g) \rbrace ,
\end{split}
\end{equation}
where $\mbox{\#}(x,g)$ is a function that returns the number of occurrences of n-gram $g$ in $x$, and $n$ is the length of n-grams. While being a rather shallow approach, string kernels attained strong results in some specific tasks. For instance, string kernels ranked first in the Arabic Dialect Identification tasks of VarDial 2017 \cite{Ionescu-VarDial-2017} and VarDial 2018 \cite{Butnaru-VarDial-2018}.

\noindent
{\bf Bag-of-word-embeddings.}
Following the seminal paper of \newcite{Mikolov-NIPS-2013} introducing \emph{word2vec}, word embeddings became one of the mainstream approaches in various computational linguistics tasks \cite{Cheng-IJCAI-2018,Conneau-EMNLP-2017,Cozma-ACL-2018,Fu-ESA-2018,Kim-EMNLP-2014,Kiros-NIPS-2015,Shen-ACL-2018,Torki-ACL-2018,Zhou-IJCAI-2018}. In order to build the bag-of-word-embeddings (BOWE), we first trained \emph{word2vec} on the collected Romanian reviews using the continuous bag-of-words (CBOW) model. Before training, we transformed all letters to lowercase and removed punctuation. In addition to word2vec, we consider a recently introduced Romanian BERT model \cite{Dumitrescu-EMNLP-2020} as an alternative way to produce word embeddings, which is likely to produce much better results, considering the success of BERT \cite{Devlin-NAACL-2019} in English NLP tasks. Instead of averaging the word embeddings to obtain document-level representations \cite{Shen-ACL-2018}, we follow a different and more effective path suggested by some recent works \cite{Ionescu-KES-2017,Cozma-ACL-2018,Fu-ESA-2018,Ionescu-NAACL-2019}. More specifically, we cluster the word embeddings collected from the entire training set using k-means. For a document $D$ of $n$ words, $D=(w_1, w_2, ..., w_n)$, a word embedding model, be it word2vec or BERT, outputs a matrix of $n\times m$ components (or a set of $n$ $m$-dimensional vectors), the $m$-dimensional vector at index $i$ corresponding to word $w_i$. We apply clustering on the word vectors extracted from all training documents, thus obtaining a set of $k$ clusters. A document $D$ is then represented as a bag-of-word-embeddings (histogram) $H=(h_1, h_2, ...., h_k)$ in which each component $h_i$ retains the number of word embeddings from the document $D$ that fall in cluster $i$, where $i \in \{1,2,...,k\}$. We note that the size of the bag-of-word-embeddings is equal to the number of clusters $k$. In the case of BERT, we emphasize that, although the embedding vector of a word depends on the context, it is likely that the embedding vectors corresponding to a specific word will fall in the same cluster. Hence, BOWE is able to cope well with this situation.

\noindent
{\bf Replacing k-means with SOMs.}
Quantitative linguistics studies \cite{Powers-CoNLL-1998} have pointed out that, given a corpus of text documents, the frequency of any word is inversely proportional to its rank in the frequency table, giving rise to a Zipf's law distribution of words in natural language. However, the k-means algorithm tends to ignore the data density, producing equally-sized clusters. We therefore propose to replace the k-means algorithm with an approach that takes into account the density in the word embedding space, producing a set of clusters that follow the Zipf's law. We propose to perform clustering using self-organizing maps (SOMs) \cite{Kohonen-Springer-2001}, since these models are known to preserve the topological properties of the input space. Indeed, Figure \ref{fig2} shows that SOMs produce clusters of Romanian word embeddings closer to the Zipf's law distribution than k-means.

It is important to emphasize that k-means can produce clusters of different size, as shown in Figure \ref{fig2}. Our observation refers only to the fact that the data density is not particularly modeled by the k-means optimization process, while SOMs are optimized by shifting the neural units following the density of the data (units tend to migrate where the space is more dense). Our observation with respect to k-means is also confirmed by other studies. For example, \newcite{Raykov-PO-2016} note that: \emph{``even when all other implicit geometric assumptions of k-means are satisfied, it will fail to learn a correct, or even meaningful, clustering when there are significant differences in cluster density''}. Since natural language involves such significant differences (due to the presence of the Zipf’s law), we believe that k-means is a sub-optimal choice.

\section{Experiments}

\begin{figure}[!t]
\begin{center}
\includegraphics[width=1.0\linewidth]{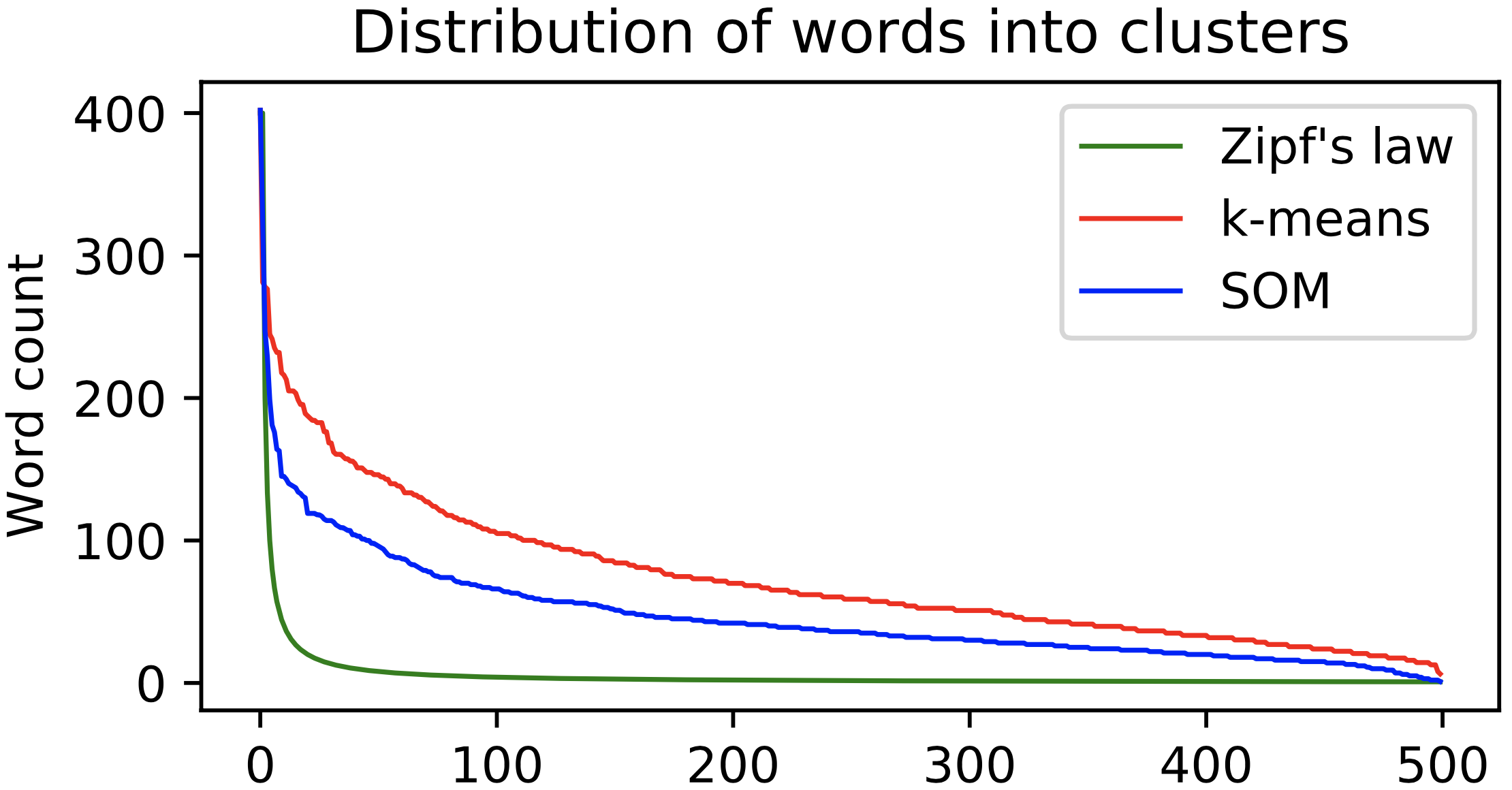}
\end{center}
\vspace*{-0.3cm}
\caption{The distribution of words into clusters generated by k-means (in red) or by SOMs (in blue), for LaRoSeDa. The Zipf's law distribution (in green) is included for reference. Best viewed in color.}
\label{fig2}
\vspace*{-0.4cm}
\end{figure}

\noindent
{\bf Corpora.}
First and foremost, we perform experiments on LaRoSeDa with the goal of introducing some benchmark results on our new data set. We also perform experiments on MOROCO \cite{Butnaru-ACL-2019}, a data set with Moldavian and Romanian news articles, with the goal of showing the generalization capacity of using SOMs instead ok k-means.

\noindent
{\bf Experimental setup.}
On LaRoSeDa, we present two sets of results, one based on the established train-test split and one based on 10-fold cross-validation. On MOROCO, we choose to present 10-fold cross-validation results for the intra-dialect multi-class categorization by topic task, on the 18,161 samples written in the Romanian dialect.

\begin{table*}[!t]
\begin{center}
\begin{tabular}{|l|c|c|}
\hline
Method 						& 10-fold CV   & Test\\
\hline
\hline
HISK				            & 84.38\%       & 84.73\% \\
BOWE-word2vec with k-means	    & 72.24\%		& 70.73\% \\
BOWE-word2vec with SOMs         & 75.99\%	    & 75.57\% \\
BOWE-BERT with k-means		    & 78.63\%		& 77.36\% \\
BOWE-BERT with SOMs             & 79.75\%	    & 80.73\% \\
HISK+BOWE-word2vec with k-means	& 85.08\%		& 83.57\% \\           
HISK+BOWE-word2vec with SOMs	& 87.17\%		& 88.93\% \\
HISK+BOWE-BERT with k-means	    & 88.81\%		& 89.42\% \\           
HISK+BOWE-BERT with SOMs	    & 89.54\%		& 90.90\% \\
\hline
\end{tabular}
\end{center}
\vspace*{-0.2cm}
\caption{Accuracy rates of HISK, BOWE-word2vec and BOWE-BERT with clustering based on k-means or SOMs, as well as ensemble models on LaRoSeDa. Results are reported in two cases: using a 10-fold cross-validation procedure and using the train-test split.}
\label{tab_results_LaRoSeDa}
\vspace*{-0.2cm}
\end{table*}

\noindent
{\bf Parameter and model choices.}
For HISK, we combined character 3-grams, 4-grams and 5-grams. For BOWE-word2vec and BOWE-BERT, we set the number of clusters to $k = 500$, just as \newcite{Cozma-ACL-2018}. We trained \emph{word2vec} to produce $300$-dimensional Romanian word embeddings, while the Romanian BERT outputs $768$-dimensional embeddings. In the learning stage, we employed the linear Support Vector Machines (SVM) implementation from Scikit-learn \cite{Pedregosa-JMLR-2011}, providing as input pre-computed kernels. For BOWE-word2vec and BOWE-BERT, we opt for the PQ kernel, based on the findings of \newcite{Ionescu-KES-2017}. We set the regularization parameter of SVM to $C=10^3$ in all the experiments. We also fuse HISK with BOWE-word2vec or BOWE-BERT in the dual form by summing up the corresponding kernel matrices. We employed an open source implementation of SOMs\footnote{http://neupy.com/pages/home.html}. We used the default choices for most hyperparameters, the modifications being detailed next. We set the learning rate to $0.25$ and the number of epochs to $200$. Before starting the training, the SOM is configured to randomly choose a number of training samples equal to number of expected outputs. We opted for the cosine distance between data samples and SOM's weights.

\noindent
{\bf Results on LaRoSeDa.}
In Table~\ref{tab_results_LaRoSeDa}, we present the results on LaRoSeDa. Among the individual baselines, we observe that HISK attains the best accuracy rates, surpassing all BOWE configurations. We also note that by replacing k-means with SOMs, the accuracy rate of BOWE-word2vec grows by 4 or 5\%. The improvements brought by SOMs can be explained by the fact that, unlike k-means, SOMs produce clusters that are closer to the Zipf's law distribution. This is proven by the word embedding counts per cluster illustrated in Figure~\ref{fig2}. When we combine HISK with BOWE-BERT, we notice significant performance gains. 

\begin{table}[!t]
\setlength\tabcolsep{1.5pt}
\begin{center}
\begin{tabular}{|l|c|}
\hline
Method 						& 10-fold CV \\
\hline
\hline
HISK \cite{Butnaru-ACL-2019} & 71.27\% \\
BOWE-BERT with k-means		& 63.42\% \\
BOWE-BERT with SOMs         & 68.50\% \\
HISK+BOWE-BERT with k-means  & 72.21\% \\           
HISK+BOWE-BERT with SOMs	 & 73.35\% \\
\hline
\end{tabular}
\end{center}
\vspace*{-0.2cm}
\caption{Accuracy rates of HISK, BOWE-BERT with k-means, BOWE-BERT with SOMs and their combinations on MOROCO, for the Romanian intra-dialect multi-class categorization by topic task. Results are reported using a 10-fold cross-validation procedure.}
\label{tab_results_MOROCO}
\vspace*{-0.3cm}
\end{table}

\noindent
{\bf Results on MOROCO.}
In Table~\ref{tab_results_MOROCO}, we present the results on MOROCO, for the Romanian intra-dialect multi-class categorization by topic task. We notice that HISK attains better results than BOWE-BERT with k-means and BOWE-BERT with SOMs, although the differences in terms of accuracy seem to be smaller. 
As for LaRoSeDa, we observe a significant improvement (higher than 5\%) when k-means is replaced by SOMs. There is an observable improvement over the plain HISK, when HISK is combined with BOWE-BERT based on k-means. Nonetheless, we notice a larger improvement when we combine HISK and BOWE-BERT based on SOMs.

\section{Conclusion}

In this paper, $(i)$ we introduced LaRoSeDa, a large data set for polarity classification of Romanian reviews, and $(ii)$ we employed self-organizing maps, a clustering approach that preserves the density of words in the embedding space, resulting in a more effective bag-of-word-embeddings representation. Our top accuracy rates on LaRoSeDa are 89.54\% for the cross-validation procedure and 90.90\% on the test set. We note that SOMs had a significant contribution in attaining these high accuracy rates. We conclude that the combination of HISK and BOWE-BERT with SOMs is a strong baseline which should encourage future research in proposing non-trivial models for Romanian polarity classification. Furthermore, the results obtained on MOROCO confirm that SOMs provide better accuracy rates than k-means when it comes to building document-level representations based on clustering word embeddings.

\section*{Acknowledgments}
The authors thank reviewers for their useful remarks. This work was supported by a grant of the Romanian Ministry of Education and Research, CNCS - UEFISCDI, project number PN-III-P1-1.1-TE-2019-0235, within PNCDI III. This article has also benefited from the support of the Romanian Young Academy, which is funded by Stiftung Mercator and the Alexander von Humboldt Foundation for the period 2020-2022.

\bibliography{refs}

\begin{thebibliography}{57}
\expandafter\ifx\csname natexlab\endcsname\relax\def\natexlab#1{#1}\fi

\bibitem[{Al-Ayyoub et~al.(2018)Al-Ayyoub, Nuseir, Alsmearat, Jararweh, and
  Gupta}]{Al-Ayyoub-JCS-2018}
Mahmoud Al-Ayyoub, Aya Nuseir, Kholoud Alsmearat, Yaser Jararweh, and Brij
  Gupta. 2018.
\newblock {Deep learning for Arabic NLP: A survey}.
\newblock \emph{Journal of Computational Science}, 26:522--531.

\bibitem[{Banea et~al.(2011)Banea, Mihalcea, and Wiebe}]{Banea-BC-2011}
Carmen Banea, Rada Mihalcea, and Janyce Wiebe. 2011.
\newblock {Multilingual Sentiment and Subjectivity Analysis}.
\newblock In \emph{Multilingual Natural Language Processing Applications: From
  Theory to Practice}, chapter~7. IBM Press.

\bibitem[{Banea et~al.(2008)Banea, Mihalcea, Wiebe, and
  Hassan}]{Banea-EMNLP-2008}
Carmen Banea, Rada Mihalcea, Janyce Wiebe, and Samer Hassan. 2008.
\newblock {Multilingual Subjectivity Analysis Using Machine Translation}.
\newblock In \emph{Proceedings of EMNLP}, page 127–135.

\bibitem[{Blitzer et~al.(2007)Blitzer, Dredze, and Pereira}]{Blitzer-ACL-2007}
John Blitzer, Mark Dredze, and Fernando Pereira. 2007.
\newblock Biographies, bollywood, boomboxes and blenders: Domain adaptation for
  sentiment classification.
\newblock In \emph{Proceedings of ACL}, pages 187--205.

\bibitem[{Briciu and Lupea(2017)}]{Briciu-SUBBI-2017}
Anamaria Briciu and Mihaiela Lupea. 2017.
\newblock {RoEmoLex -- A Romanian Emotion Lexicon}.
\newblock \emph{Studia Universitatis Babeș-Bolyai Informatica}, 62(2):45--56.

\bibitem[{Brooke et~al.(2009)Brooke, Tofiloski, and
  Taboada}]{Brooke-RANLP-2009}
Julian Brooke, Milan Tofiloski, and Maite Taboada. 2009.
\newblock {Cross-linguistic sentiment analysis: From English to Spanish}.
\newblock In \emph{Proceedings of RANLP}, pages 50--54.

\bibitem[{Butnaru and Ionescu(2017)}]{Ionescu-KES-2017}
Andrei Butnaru and Radu~Tudor Ionescu. 2017.
\newblock {From Image to Text Classification: A Novel Approach based on
  Clustering Word Embeddings}.
\newblock In \emph{Proceedings of KES}, pages 1784--1793.

\bibitem[{Butnaru and Ionescu(2019)}]{Butnaru-ACL-2019}
Andrei Butnaru and Radu~Tudor Ionescu. 2019.
\newblock {MOROCO: The Moldavian and Romanian Dialectal Corpus}.
\newblock In \emph{Proceedings of ACL}, pages 688--698.

\bibitem[{Butnaru and Ionescu(2018)}]{Butnaru-VarDial-2018}
Andrei~M. Butnaru and Radu~Tudor Ionescu. 2018.
\newblock {UnibucKernel Reloaded: First Place in Arabic Dialect Identification
  for the Second Year in a Row}.
\newblock In \emph{Proceedings of VarDial}, pages 77--87.

\bibitem[{Cheng et~al.(2018)Cheng, Yuan, Li, and Yang}]{Cheng-IJCAI-2018}
Zhou Cheng, Chun Yuan, Jiancheng Li, and Haiqin Yang. 2018.
\newblock {TreeNet: Learning Sentence Representations with Unconstrained Tree
  Structure}.
\newblock In \emph{Proceedings of IJCAI}, pages 4005--4011.

\bibitem[{Colhon et~al.(2016)Colhon, Cerban, Becheru, and
  Teodorescu}]{Colhon-INISTA-2016}
Mihaela Colhon, M{\u{a}}d{\u{a}}lina Cerban, Alex Becheru, and Mirela
  Teodorescu. 2016.
\newblock {Polarity shifting for Romanian sentiment classification}.
\newblock In \emph{Proceedings of INISTA}, pages 1--6.

\bibitem[{Conneau et~al.(2017)Conneau, Kiela, Schwenk, Barrault, and
  Bordes}]{Conneau-EMNLP-2017}
Alexis Conneau, Douwe Kiela, Holger Schwenk, Lo{\"\i}c Barrault, and Antoine
  Bordes. 2017.
\newblock {Supervised Learning of Universal Sentence Representations from
  Natural Language Inference Data}.
\newblock In \emph{Proceedings of EMNLP}, pages 670--680.

\bibitem[{Cozma et~al.(2018)Cozma, Butnaru, and Ionescu}]{Cozma-ACL-2018}
M\u{a}d\u{a}lina Cozma, Andrei Butnaru, and Radu~Tudor Ionescu. 2018.
\newblock Automated essay scoring with string kernels and word embeddings.
\newblock In \emph{Proceedings of ACL}, pages 503--509.

\bibitem[{Dahou et~al.(2016)Dahou, Xiong, Zhou, Haddoud, and
  Duan}]{Dahou-COLING-2016}
Abdelghani Dahou, Shengwu Xiong, Junwei Zhou, Mohamed~Houcine Haddoud, and
  Pengfei Duan. 2016.
\newblock {Word embeddings and convolutional neural network for Arabic
  sentiment classification}.
\newblock In \emph{Proceedings of COLING}, pages 2418--2427.

\bibitem[{Devlin et~al.(2019)Devlin, Chang, Lee, and
  Toutanova}]{Devlin-NAACL-2019}
Jacob Devlin, Ming-Wei Chang, Kenton Lee, and Kristina Toutanova. 2019.
\newblock {BERT: Pre-training of Deep Bidirectional Transformers for Language
  Understanding}.
\newblock In \emph{Proceedings of NAACL}, pages 4171--4186.

\bibitem[{Dos~Santos and Gatti(2014)}]{Dos-COLING-2014}
C{\'\i}cero~Nogueira Dos~Santos and Maira Gatti. 2014.
\newblock {Deep Convolutional Neural Networks for Sentiment Analysis of Short
  Texts}.
\newblock In \emph{Proceedings of COLING}, pages 69--78.

\bibitem[{Dumitrescu et~al.(2020)Dumitrescu, Avram, and
  Pyysalo}]{Dumitrescu-EMNLP-2020}
{\c{S}tefan Daniel}~Dumitrescu, Andrei-Marius Avram, and Sampo Pyysalo. 2020.
\newblock {The birth of Romanian BERT}.
\newblock In \emph{Findings of EMNLP}.

\bibitem[{Elsahar and El-Beltagy(2015)}]{Elsahar-CICLing-2015}
Hady Elsahar and Samhaa~R. El-Beltagy. 2015.
\newblock {Building large Arabic multi-domain resources for sentiment
  analysis}.
\newblock In \emph{Proceedings of CICLing}, pages 23--34.

\bibitem[{Fu et~al.(2018)Fu, Qu, Huang, and Lu}]{Fu-ESA-2018}
Mingsheng Fu, Hong Qu, Li~Huang, and Li~Lu. 2018.
\newblock {Bag of meta-words: A novel method to represent document for the
  sentiment classification}.
\newblock \emph{Expert Systems with Applications}, 113:33--43.

\bibitem[{Gim\'{e}nez-P\'{e}rez et~al.(2017)Gim\'{e}nez-P\'{e}rez,
  Franco-Salvador, and Rosso}]{Gimenez-EACL-2017}
Rosa~M. Gim\'{e}nez-P\'{e}rez, Marc Franco-Salvador, and Paolo Rosso. 2017.
\newblock {Single and Cross-domain Polarity Classification using String
  Kernels}.
\newblock In \emph{Proceedings of EACL}, pages 558--563.

\bibitem[{G{\^\i}nsc{\u{a}} et~al.(2011)G{\^\i}nsc{\u{a}},
  Boro{\textcommabelow{s}}, Iftene, Trandab{\u{a}}{\textcommabelow{t}}, Toader,
  Cor{\^\i}ci, Perez, and Cristea}]{Ginsca-WASSA-2011}
Alexandru-Lucian G{\^\i}nsc{\u{a}}, Emanuela Boro{\textcommabelow{s}}, Adrian
  Iftene, Diana Trandab{\u{a}}{\textcommabelow{t}}, Mihai Toader, Marius
  Cor{\^\i}ci, Cenel-Augusto Perez, and Dan Cristea. 2011.
\newblock {Sentimatrix -- Multilingual Sentiment Analysis Service}.
\newblock In \emph{Proceedings of WASSA}, pages 189--195.

\bibitem[{Huang et~al.(2017)Huang, Rao, Xie, Wong, and Wang}]{Huang-AAAI-2017}
Xingchang Huang, Yanghui Rao, Haoran Xie, Tak-Lam Wong, and Fu~Lee Wang. 2017.
\newblock {Cross-Domain Sentiment Classification via Topic-Related TrAdaBoost}.
\newblock In \emph{Proceedings of AAAI}, pages 4939--4940.

\bibitem[{Ionescu and Butnaru(2019)}]{Ionescu-NAACL-2019}
Radu~Tudor Ionescu and Andrei Butnaru. 2019.
\newblock {Vector of Locally-Aggregated Word Embeddings (VLAWE): A Novel
  Document-level Representation}.
\newblock In \emph{Proceedings of NAACL}, pages 363--369.

\bibitem[{Ionescu and Butnaru(2017)}]{Ionescu-VarDial-2017}
Radu~Tudor Ionescu and Andrei~M. Butnaru. 2017.
\newblock {Learning to Identify Arabic and German Dialects using Multiple
  Kernels}.
\newblock In \emph{Proceedings of VarDial}, pages 200--209.

\bibitem[{Ionescu and Butnaru(2018)}]{Ionescu-EMNLP-2018}
Radu~Tudor Ionescu and Andrei~M. Butnaru. 2018.
\newblock {Improving the results of string kernels in sentiment analysis and
  Arabic dialect identification by adapting them to your test set}.
\newblock In \emph{Proceedings of EMNLP}, pages 1084--1090.

\bibitem[{Ionescu et~al.(2014)Ionescu, Popescu, and
  Cahill}]{Ionescu-EMNLP-2014}
Radu~Tudor Ionescu, Marius Popescu, and Aoife Cahill. 2014.
\newblock {Can characters reveal your native language? A language-independent
  approach to native language identification}.
\newblock In \emph{Proceedings of EMNLP}, pages 1363--1373.

\bibitem[{Ionescu et~al.(2016)Ionescu, Popescu, and Cahill}]{Ionescu-COLI-2016}
Radu~Tudor Ionescu, Marius Popescu, and Aoife Cahill. 2016.
\newblock String kernels for native language identification: Insights from
  behind the curtains.
\newblock \emph{Computational Linguistics}, 42(3):491--525.

\bibitem[{Kim(2014)}]{Kim-EMNLP-2014}
Yoon Kim. 2014.
\newblock {Convolutional Neural Networks for Sentence Classification}.
\newblock In \emph{Proceedings of EMNLP}, pages 1746--1751.

\bibitem[{Kiros et~al.(2015)Kiros, Zhu, Salakhutdinov, Zemel, Urtasun,
  Torralba, and Fidler}]{Kiros-NIPS-2015}
Ryan Kiros, Yukun Zhu, Ruslan~R. Salakhutdinov, Richard Zemel, Raquel Urtasun,
  Antonio Torralba, and Sanja Fidler. 2015.
\newblock {Skip-Thought Vectors}.
\newblock In \emph{Proceedings of NIPS}, pages 3294--3302.

\bibitem[{Kohonen(2001)}]{Kohonen-Springer-2001}
Teuvo Kohonen. 2001.
\newblock \emph{{Self-Organizing Maps}}, 3rd edition.
\newblock Springer-Verlag.

\bibitem[{Lupea and Briciu(2017)}]{Lupea-ICCP-2017}
Mihaiela Lupea and Anamaria Briciu. 2017.
\newblock {Formal concept analysis of a Romanian emotion lexicon}.
\newblock In \emph{Proceedings of ICCP}, pages 111--118.

\bibitem[{Maas et~al.(2011)Maas, Daly, Pham, Huang, Ng, and
  Potts}]{Maas-ACL-2011}
Andrew~L. Maas, Raymond~E. Daly, Peter~T. Pham, Dan Huang, Andrew~Y. Ng, and
  Christopher Potts. 2011.
\newblock {Learning Word Vectors for Sentiment Analysis}.
\newblock In \emph{Proceedings of ACL}, pages 142--150.

\bibitem[{Mihalcea et~al.(2007)Mihalcea, Banea, and Wiebe}]{Mihalcea-ACL-2007}
Rada Mihalcea, Carmen Banea, and Janyce Wiebe. 2007.
\newblock Learning multilingual subjective language via cross-lingual
  projections.
\newblock In \emph{Proceedings of ACL}, pages 976--983.

\bibitem[{Mikolov et~al.(2013)Mikolov, Sutskever, Chen, Corrado, and
  Dean}]{Mikolov-NIPS-2013}
Tomas Mikolov, Ilya Sutskever, Kai Chen, Gregory~S. Corrado, and Jeffrey Dean.
  2013.
\newblock {Distributed Representations of Words and Phrases and their
  Compositionality}.
\newblock In \emph{Proceedings of NIPS}, pages 3111--3119.

\bibitem[{Molina-Gonz{\'a}lez et~al.(2015)Molina-Gonz{\'a}lez,
  Mart{\'\i}nez-C{\'a}mara, Mart{\'\i}n-Valdivia, and
  Ure{\~n}a-L{\'o}pez}]{Molina-IPM-2015}
Dolores~M. Molina-Gonz{\'a}lez, Eugenio Mart{\'\i}nez-C{\'a}mara, Teresa~M.
  Mart{\'\i}n-Valdivia, and Alfonso~L. Ure{\~n}a-L{\'o}pez. 2015.
\newblock {A Spanish semantic orientation approach to domain adaptation for
  polarity classification}.
\newblock \emph{Information Processing \& Management}, 51(4):520--531.

\bibitem[{Nabil et~al.(2015)Nabil, Aly, and Atiya}]{Nabil-EMNLP-2015}
Mahmoud Nabil, Mohamed Aly, and Amir Atiya. 2015.
\newblock {ASTD: Arabic Sentiment Tweets Dataset}.
\newblock In \emph{Proceedings of EMNLP}, pages 2515--2519.

\bibitem[{Nabil et~al.(2013)Nabil, Aly, and Atiya}]{Nabil-ACL-2013}
Mahmoud Nabil, Mohamed~A. Aly, and Amir~F. Atiya. 2013.
\newblock {LABR: A Large Scale Arabic Book Reviews Dataset}.
\newblock In \emph{Proceedings of ACL}, pages 494--498.

\bibitem[{Navas-Loro and Rodr{\'\i}guez-Doncel(2019)}]{Navas-LRE-2019}
Mar{\'\i}a Navas-Loro and V{\'\i}ctor Rodr{\'\i}guez-Doncel. 2019.
\newblock Spanish corpora for sentiment analysis: a survey.
\newblock \emph{Language Resources and Evaluation}, pages 1--38.

\bibitem[{Pang and Lee(2005)}]{Pang-ACL-2005}
Bo~Pang and Lillian Lee. 2005.
\newblock {Seeing Stars: Exploiting Class Relationships For Sentiment
  Categorization With Respect To Rating Scales}.
\newblock In \emph{Proceedings of ACL}, pages 115--124.

\bibitem[{Pedregosa et~al.(2011)Pedregosa, Varoquaux, Gramfort, Michel,
  Thirion, Grisel, Blondel, Prettenhofer, Weiss, Dubourg, Vanderplas, Passos,
  Cournapeau, Brucher, Perrot, and Duchesnay}]{Pedregosa-JMLR-2011}
F.~Pedregosa, G.~Varoquaux, A.~Gramfort, V.~Michel, B.~Thirion, O.~Grisel,
  M.~Blondel, P.~Prettenhofer, R.~Weiss, V.~Dubourg, J.~Vanderplas, A.~Passos,
  D.~Cournapeau, M.~Brucher, M.~Perrot, and E.~Duchesnay. 2011.
\newblock {Scikit-learn: Machine Learning in {P}ython}.
\newblock \emph{Journal of Machine Learning Research}, 12:2825--2830.

\bibitem[{Peng et~al.(2017)Peng, Cambria, and Hussain}]{Peng-CC-2017}
Haiyun Peng, Erik Cambria, and Amir Hussain. 2017.
\newblock {A review of sentiment analysis research in Chinese language}.
\newblock \emph{Cognitive Computation}, 9(4):423--435.

\bibitem[{Popescu et~al.(2017)Popescu, Grozea, and Ionescu}]{Popescu-KES-2017}
Marius Popescu, Cristian Grozea, and Radu~Tudor Ionescu. 2017.
\newblock {HASKER: An efficient algorithm for string kernels. Application to
  polarity classification in various languages}.
\newblock In \emph{Proceedings of KES}, pages 1755--1763.

\bibitem[{Powers(1998)}]{Powers-CoNLL-1998}
David~M.W. Powers. 1998.
\newblock {Applications and Explanations of Zipf's Law}.
\newblock In \emph{Proceedings of CoNLL}, pages 151--160.

\bibitem[{Raykov et~al.(2016)Raykov, Boukouvalas, Baig, and
  Little}]{Raykov-PO-2016}
Yordan~P. Raykov, Alexis Boukouvalas, Fahd Baig, and Max~A. Little. 2016.
\newblock {What to Do When K-Means Clustering Fails: A Simple yet Principled
  Alternative Algorithm}.
\newblock \emph{PLOS ONE}, 11(9):1--28.

\bibitem[{Russu et~al.(2014)Russu, Dinsoreanu, Vlad, and
  Potolea}]{Russu-ICCP-2014}
Roxana~Monica Russu, Mihaela Dinsoreanu, Oana~Lumini\c{t}a Vlad, and Rodica
  Potolea. 2014.
\newblock {An opinion mining approach for Romanian language}.
\newblock In \emph{Proceedings of ICCP}, pages 43--46.

\bibitem[{Shen et~al.(2018)Shen, Wang, Wang, Min, Su, Zhang, Li, Henao, and
  Carin}]{Shen-ACL-2018}
Dinghan Shen, Guoyin Wang, Wenlin Wang, Martin~Renqiang Min, Qinliang Su, Yizhe
  Zhang, Chunyuan Li, Ricardo Henao, and Lawrence Carin. 2018.
\newblock {Baseline Needs More Love: On Simple Word-Embedding-Based Models and
  Associated Pooling Mechanisms}.
\newblock In \emph{Proceedings of ACL}, pages 440--450.

\bibitem[{Socher et~al.(2013)Socher, Perelygin, Wu, Chuang, Manning, Ng, and
  Potts}]{Socher-EMNLP-2013}
Richard Socher, Alex Perelygin, Jean Wu, Jason Chuang, D.~Christopher Manning,
  Andrew Ng, and Christopher Potts. 2013.
\newblock {Recursive Deep Models for Semantic Compositionality Over a Sentiment
  Treebank}.
\newblock In \emph{Proceedings of EMNLP}, pages 1631--1642.

\bibitem[{Sokolova and Bobicev(2009)}]{Sokolova-RANLP-2009}
Marina Sokolova and Victoria Bobicev. 2009.
\newblock {Classification of Emotion Words in Russian and Romanian Languages}.
\newblock In \emph{Proceedings of RANLP}, pages 416--420.

\bibitem[{Torki(2018)}]{Torki-ACL-2018}
Marwan Torki. 2018.
\newblock {A Document Descriptor using Covariance of Word Vectors}.
\newblock In \emph{Proceedings of ACL}, pages 527--532.

\bibitem[{Vilares et~al.(2015)Vilares, Alonso, and
  G{\'o}mez-Rodr{\'\i}guez}]{Vilares-NLE-2015}
David Vilares, Miguel~A. Alonso, and Carlos G{\'o}mez-Rodr{\'\i}guez. 2015.
\newblock {A syntactic approach for opinion mining on Spanish reviews}.
\newblock \emph{Natural Language Engineering}, 21(1):139--163.

\bibitem[{Wan(2008)}]{Wan-EMNLP-2008}
Xiaojun Wan. 2008.
\newblock {Using Bilingual Knowledge and Ensemble Techniques for Unsupervised
  Chinese Sentiment Analysis}.
\newblock In \emph{Proceedings of EMNLP}, pages 553--561.

\bibitem[{Zafra et~al.(2017)Zafra, Mart{\'\i}n-Valdivia, Camara, and
  Ure{\~n}a-L{\'o}pez}]{Zafra-TAC-2017}
Salud Maria~Jimenez Zafra, Teresa~M. Mart{\'\i}n-Valdivia, Eugenio~Martinez
  Camara, and Alfonso~L. Ure{\~n}a-L{\'o}pez. 2017.
\newblock {Studying the scope of negation for Spanish sentiment analysis on
  Twitter}.
\newblock \emph{IEEE Transactions on Affective Computing}, 10(1):129--141.

\bibitem[{Zagibalov and Carroll(2008)}]{Zagibalov-IJCNLP-2008}
Taras Zagibalov and John Carroll. 2008.
\newblock {Unsupervised classification of sentiment and objectivity in Chinese
  text}.
\newblock In \emph{Proceedings of IJCNLP}, pages 304--311.

\bibitem[{Zhai et~al.(2011)Zhai, Xu, Kang, and Jia}]{Zhai-ESA-2011}
Zhongwu Zhai, Hua Xu, Bada Kang, and Peifa Jia. 2011.
\newblock {Exploiting effective features for Chinese sentiment classification}.
\newblock \emph{Expert Systems with Applications}, 38(8):9139--9146.

\bibitem[{Zhang et~al.(2008)Zhang, Zuo, Peng, and He}]{Zhang-ICCIT-2008}
Changli Zhang, Wanli Zuo, Tao Peng, and Fengling He. 2008.
\newblock {Sentiment Classification for Chinese Reviews Using Machine Learning
  Methods Based on String Kernel}.
\newblock In \emph{Proceedings of ICCIT}, volume~2, pages 909--914.

\bibitem[{Zhang and He(2013)}]{Zhang-JIS-2013}
Pu~Zhang and Zhongshi He. 2013.
\newblock {A weakly supervised approach to Chinese sentiment classification
  using partitioned self-training}.
\newblock \emph{Journal of Information Science}, 39(6):815--831.

\bibitem[{Zhou et~al.(2018)Zhou, Wang, and Dong}]{Zhou-IJCAI-2018}
Qianrong Zhou, Xiaojie Wang, and Xuan Dong. 2018.
\newblock Differentiated attentive representation learning for sentence
  classification.
\newblock In \emph{Proceedings of IJCAI}, pages 4630--4636.

\end{thebibliography}
\bibliographystyle{acl_natbib}

\end{document}